\documentclass[a4paper]{article}

\usepackage{INTERSPEECH2019}

\usepackage{kotex}
\usepackage{xcolor}
\usepackage{multirow}
\usepackage{subcaption}
\usepackage{enumitem}

\title{Effective Sentence Scoring Method using Bidirectional Language Model\\ for Speech Recognition}
\name{Joongbo Shin, Yoonhyung Lee, and Kyomin Jung}
\address{
  Seoul National University\\
  Republic of Korea}
\email{\{jbshin,cpi1234,kjung\}@snu.ac.kr}

\begin{document}

\maketitle
\begin{abstract}
In automatic speech recognition, many studies have shown performance improvements using language models (LMs).
Recent studies have tried to use bidirectional LMs (biLMs) instead of conventional unidirectional LMs (uniLMs) for rescoring the $N$-best list decoded from the acoustic model.
In spite of their theoretical benefits, the biLMs have not given notable improvements compared to the uniLMs in their experiments. 
This is because their biLMs do not consider the interaction between the two directions.
In this paper, we propose a novel sentence scoring method considering the interaction between the past and the future words on the biLM.
Our experimental results on the LibriSpeech corpus show that the biLM with the proposed sentence scoring outperforms the uniLM for the $N$-best list rescoring, consistently and significantly in all experimental conditions.
The analysis of WERs by word position demonstrates that the biLM is more robust than the uniLM especially when a recognized sentence is short or a misrecognized word is at the beginning of the sentence.

\end{abstract}

\section{Introduction}
A language model (LM) is an essential component in recent automatic speech recognition (ASR) systems.
Since the LM captures the possibility of any word sequence, it can help to distinguish between words with similar sounds.
Conventionally, LMs are used to predict the probability of the next word given its preceding words. 
Many state-of-the-art speech recognition systems have achieved performance improvements with these \textit{unidirectional} LMs (uniLMs), including $n$-gram LMs \cite{heafield2013scalable} and recurrent neural network LMs \cite{mikolov2010recurrent}.

Recently, \textit{bidirectional} LMs (biLMs) have achieved significant success in many applications of the natural language processing \cite{peters2018deep,devlin2018bert,}.
In speech recognition, there have also been several studies that use biLMs to capture the full context rather than just the previous words \cite{arisoy2015bidirectional,chen2017investigating}.
Even though bidirectional networks are superior to unidirectional ones in many applications from phoneme classification \cite{graves2005framewise} to acoustic modeling \cite{chan2016listen}, 
the biLMs for ASR did not show their excellence compared to the uniLMs when applying the LMs to the rescoring. 
This is because there is no interaction between the \textit{past} and \textit{future} words in their biLMs, although the words on both sides are used to predict the current word.
Namely, the forward and backward representations are not be fused in their models, and it may limit the biLM's potential. 
Furthermore, because their biLM architectures are restricted on one encoding layer, the capacities of the LMs are insufficient to model complex patterns of the human language. 
Similar issues have been addressed in recent studies on pre-training the language representation models \cite{peters2018deep,devlin2018bert}. 

In this paper, we propose a novel sentence scoring method which reflects the interactions between the past and the future words on the biLM.
The main idea of our sentence scoring is replacing a word with a special token \textless M\textgreater~in a given sentence, and then making the biLM predict the original word of the masked position using only its surrounding words.
In our model, the past and the future representations interact each other during computing the probability of the masked word.
It is important to make the prediction task non-trivial by hiding the meaning of the word while leaving only its positional information, otherwise the model simply copies the exposed words.
The score of a sentence is obtained by computing the probability of each word at a time in the given sentence, and then aggregating all the probabilities.
This sentence score is used to rescore the $N$-best list of the speech recognition outputs, following the previous studies on biLMs for ASR \cite{arisoy2015bidirectional,chen2017investigating}.

Experiments on the 1000-hour LibriSpeech ASR task \cite{panayotov2015librispeech} demonstrate that the proposed scoring method with the biLM is considerably better than the traditional scoring method with the uniLM for rescoring the $N$-best sentence list.
Our biLM achieves a 22.2\% relative improvement in word error rate (WER) to the baseline recognition system on the test-clean set, while the uniLM shows a 16.3\% relative improvement.
Moreover, the additional analysis of WERs by word position shows that the biLM is more robust than the uniLM especially when a recognized sentence is short or a misrecognized word is at the earlier part of the sentence. 
To the best of our knowledge, this is the first study that the biLM significantly and consistently outperforms the uniLM for ASR.

The rest of this paper is organized as follows: 
Section 2 introduces our sentence scoring method including the biLM and uniLM we use here and how to rescore the $N$-best list using the LMs.
Section 3 describes experimental setups for the baseline recognition system and the language model parameters.
We present the results of the experiment and its discussion in Section 4, and draw the conclusion in Section 5.

\section{Methodology}
\label{sec:method}
In this section, we present the sentence scoring method which uses bidirectional language model (LM) for rescoring the $N$-best list.
First, we review the key operations of the self-attention network (SAN), which is the base architecture of our LMs.
We then outline the architecture of the bidirectional SANLM (biSANLM) and how to train it.
In Section 2.3, we introduce the procedure of the sentence scoring, which is the core development in this paper.
Section 2.4 describes the unidirectional LM consisting of the SAN (uniSANLM), which is used as a comparison model.

\subsection{Self-Attention Network}
\label{sec:san}
\begin{figure}[t]
    \centering
    \includegraphics[width=.8\linewidth]{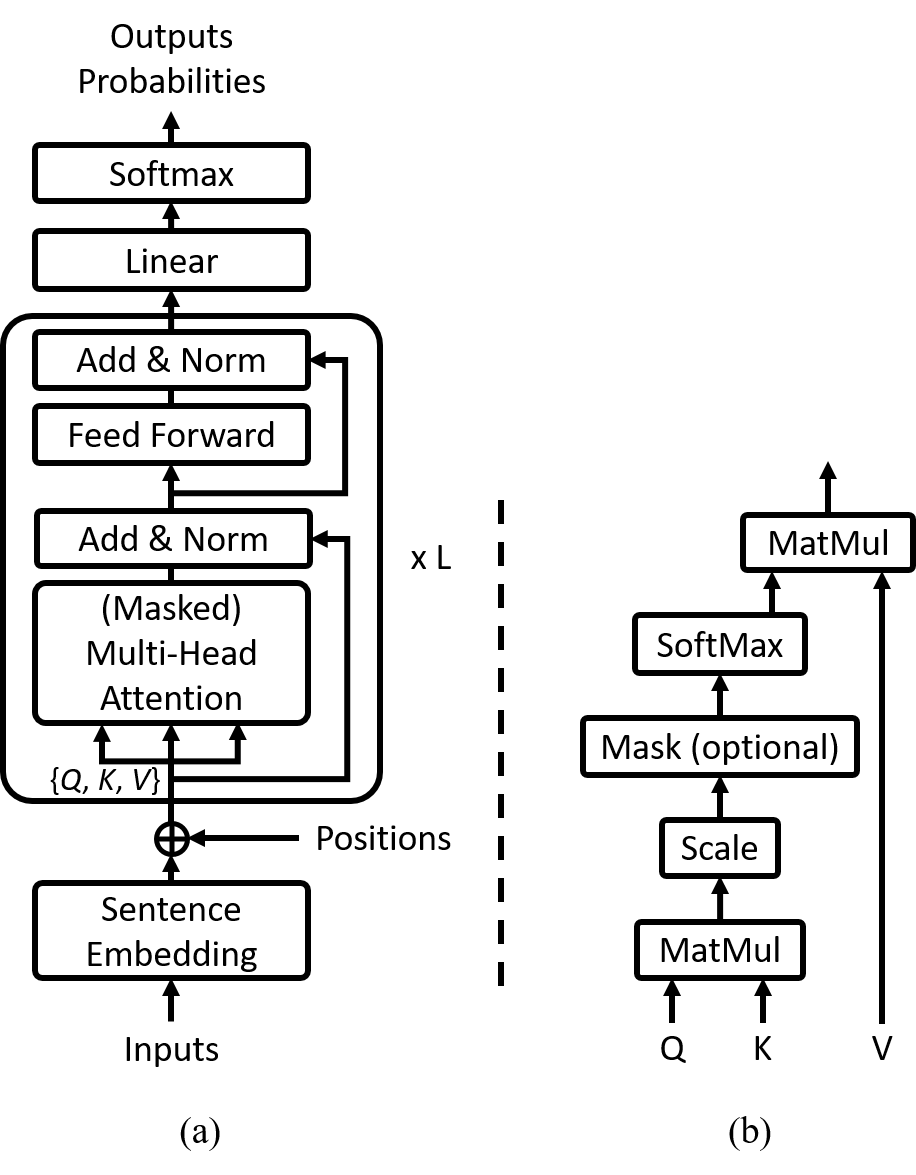}
    \caption{Architectures of (a) the self attention network and (b) the scaled dot-product attention.}
    \label{fig:san}
\end{figure}

This section briefly overviews the key operations of the self-attention network (SAN) known as the Transformer \cite{vaswani2017attention}.
While recurrent neural networks (RNNs) appear to be a natural choice for language modeling, SANs have recently shown competitive performance on sequence modeling with a slight trade-off between speed and accuracy \cite{peters2018dissecting,tang2018self}.
Note that the main interest in this paper is the comparison between the biLM and the uniLM, and we choose the SAN for the LMs' common architecture, primarily for speed.

Self-attention, often called intra-attention, is an attention mechanism that computes the representation of a single sequence by relating all positions by themselves.
This computation can be done by using the scaled dot-product attention:
\begin{equation}
    \text{Attention}(Q, K, V) = \text{Softmax}(\frac{QK^T}{\sqrt{d_k}})V,
\end{equation}
where $Q, K, V$ are query, key, value matrices respectively, which are generated from the input sequence $X\in\mathbb{R}^{n\times d}$ with the number of words $n$, the input dimension $d$.
To make the model aware of the word position in a sentence, we use the position embedding that is added to the sentence matrix of the input.
To leverage the capacity of the SAN, multi-head self-attention is applied:
\begin{equation}
\begin{split}
    \text{MultiHead}(Q, K, V) = \text{Concat}(\text{head}_1,\ldots,\text{head}_h)W^O,\\
    \text{where}~\text{head}_i=\text{Attention}(QW_i^Q, KW_i^K, VW_i^V),
\end{split}
\end{equation}
$W_i^Q, W_i^K, W_i^V\in\mathbb{R}^{d\times d_k}$ and $W^O\in\mathbb{R}^{d_{k}h\times d}$ are the parameter matrices for projections with the number of heads $h$, and $d_k=d/h$ is used for reducing the computational cost.
In addition, the position-wise feed-forward network, the layer normalization, residual connection, and dropout are also used in the SAN module for effective training.
Figure \ref{fig:san} shows the self-attention network with the scaled dot-product attention, and detail formulas are the same as in the original Transformer \cite{vaswani2017attention}.

\subsection{Bidirectional SANLM}
\label{sec:bilm}
\begin{figure}[t]
    \centering
    \includegraphics[width=.9\linewidth]{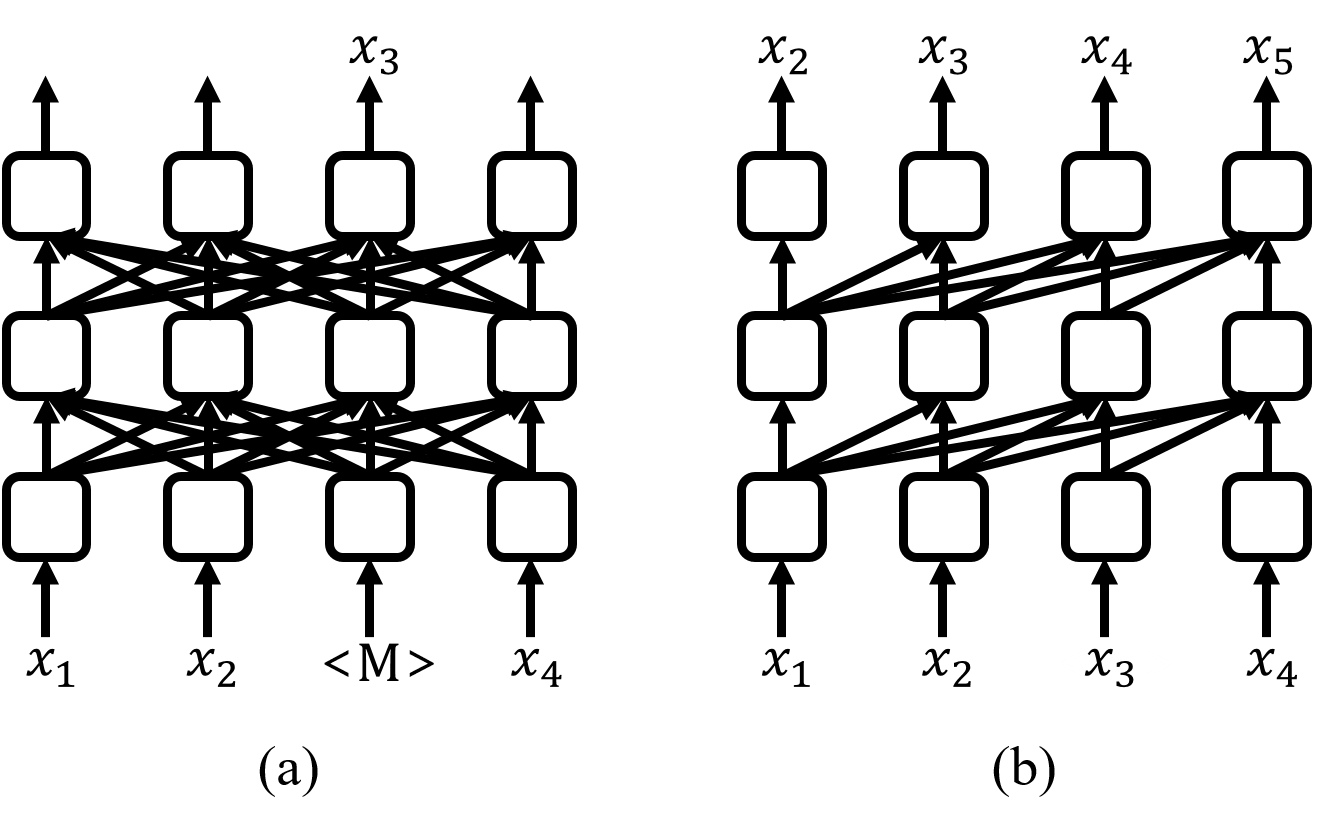}
    \caption{Schematic diagrams of (a) bidirectional SANLM for masked word prediction and (b) unidirectional SANLM for next word prediction.}
    \label{fig:lms}
\end{figure}
We now explain the architecture of the biSANLM, which is used for scoring a given sentence in the next section. 
As shown in Figure \ref{fig:san}a, our biSANLM architecture is similar to the encoder of the Transformer \cite{vaswani2017attention} except for having more layers such as the softmax layer with the linear projection.
Let $X_S\in \mathbb{R}^{n\times d}$ be a sentence matrix which is the input embeddings of the LM.
In order to make the model aware of the order of the words in the sequence, we add absolute position embeddings $X_P\in \mathbb{R}^{n\times d}$ to the input embeddings at the bottom of the encoder, and thus we have $X=X_S+X_P$ as the input of the LM.
On top of this input, we can build an encoder by stacking the SAN layers as many as we want.
The output sequence of the highest SAN, $Y_L\in \mathbb{R}^{n\times d}$ with the number of layers $L$, is used to predict the probabilities of the words through the softmax layer with the linear projection.

To train our biSANLM, we consider the masked language modeling (MLM) objective of the BERT \cite{devlin2018bert}. 
Specifically, the biSANLM learns to predict the original word in the masked position in the input sentence, which is also known as the Cloze task \cite{taylor1953cloze}.
However, our training approach has many differences from that of the BERT because our purpose of training bidirectional LM is for scoring a sentence rather than for fine-tuning the model to the other task \cite{devlin2018bert}. 
First, we sample randomly 15\% of the words from the sentence as in the BERT, but replace them by \textless M\textgreater~tokens \textit{all the time} unlike in the BERT \cite{devlin2018bert}.
Second, our training instance has a single sentence (maximum 128 words) instead of multiple sentences.
Lastly, the maximum number of masked tokens in a training instance is limited by the small number 4, because our instance has only one sentence and too much loss in information is unhelpful to train the model.
Note that we make the training instances have multiple masked tokens \textless M\textgreater~for efficient training, while we make the inference instance have only one \textless M\textgreater~for the sentence scoring.

\subsection{Our Sentence Scoring Method}
\label{sec:scoring}
This section introduces our sentence scoring method, the core development in this paper.
The basic principle of our sentence scoring method is to mask one word in a given sentence and then compute the probability of the original word on the masked position using the biSANLM.
Because the whole sentence with the masked word is taken to the biSANLM as an input, both past and future representations can be fused without making the task trivial.

Our sentence scoring method follows the procedure:
First, we create a set of instances from a given sentence by replacing each word with the predefined token \textless M\textgreater~at a time.
For example, if the sentence has seven words, we create seven instances as below:
\begin{itemize}[leftmargin=*]
\item \textbf{A given sentence}:
    \item[] \texttt{move the vat over the hot fire}
\item \textbf{A set of instances we create}:
    \begin{itemize}
    \item[1.] Input $=$ \texttt{\textless M\textgreater~the vat over the hot fire}
    \item[] Label $=$ \texttt{move}
    \item[2.] Input $=$ \texttt{move \textless M\textgreater~vat over the hot fire}
    \item[] Label $=$ \texttt{the}
    \item[] $\cdots$
    \item[7.] Input $=$ \texttt{move the vat over the hot \textless M\textgreater}
    \item[] Label $=$ \texttt{fire}
    \end{itemize}
\end{itemize}
After the creation, our biSANLM takes each instance and computes the probability of the original word in the masked position as shown in Figure \ref{fig:lms}a.
By aggregating all probabilities of the words from all instances, the score of the given sentence is obtained.

In this work, we consider the sum of all log-likelihoods of a masked word in each input sentence as the sentence score of the biSANLM.
Following the previous works on bidirectional language models for speech recognition \cite{arisoy2015bidirectional,chen2017investigating}, we use our sentence score for rescoring the $N$-best hypotheses.
For simplicity, we linearly interpolate the scores obtained by the acoustic model (AM) and the language model (LM):
\begin{equation}
\label{eq:inter2}
    \text{score}=(1-\lambda)\cdot\text{score}_{\text{AM}}+\lambda\cdot\text{score}_{\text{LM}},
\end{equation}
and $\lambda$ is the interpolation weight, which is determined empirically on development data. 
Although it is not straightforward to compare the perplexity of the biSANLM with that of the traditional (unidirectional) LM, the log-likelihood of the masked word can still be used as the score for the $N$-best list rescoring.

\subsection{Unidirectional SANLM}
This section outlines the unidirectional SANLM (uniSANLM), which is a comparison model for our biSANLM.
For a fair comparison, our uniSANLM and biSANLM have almost the same architecture, including the summation of sentence embeddings and position embeddings, the number of SAN layers, and the softmax layer.
However, the uniLM has one additional operation of masking the key-query attention, which is an optional operation in the scaled dot-product attention shown in Figure \ref{fig:san}b.
This masking operation prevents words from attending to the future words, by making the upper triangle of the key-query attention to be 0 as in the decoder of the Transformer \cite{vaswani2017attention}.
The uniSANLM is trained with the next word prediction task as in the traditional LMs.
Figure \ref{fig:lms}b shows an example of the uniSANLM that predicts next word using only its preceding words.
Following the scoring method in Section 2.3, we consider the sum of all log-likelihoods of the next word in an input sentence as the sentence score of the uniSANLM.
We also use Equation \ref{eq:inter2} to combine the AM with the uniSANLM.

\section{Experimental Setups}
We evaluate the proposed approach on the LibriSpeech ASR task \cite{panayotov2015librispeech}.
The 960-hour of training data is used to train an acoustic model, which is our baseline recognition system.
We obtain the 100-best hypothesis list for each audio in development and test data using the acoustic model, and then use the biSANLM and the uniSANLM to rescore these 100-best lists, following the previous studies on biLMs for ASR \cite{arisoy2015bidirectional,chen2017investigating}.
The details of the baseline acoustic model and language model settings are explained in the following sections.

\subsection{Baseline Acoustic Model}
In this study, we use the attention-based seq2seq model \textit{Listen, Attend and Spell} (LAS) \cite{chan2016listen} as our baseline acoustic model with some differences.
First, there are additional bottleneck fully connected (FC) layers between every bidirectional long-short term memory (BLSTM) layer.
Second, the number of time steps is reduced in half by just subsampling hidden states for even number time steps before the FC layer, instead of concatenating every two hidden states.
Third, LAS is trained with additional CTC objective function because the left-to-right constraint of CTC helps LAS learn alignments between speech-text pairs \cite{hori2017advances}.

The details of our acoustic model follow the default settings provided in ESPNet toolkit v.0.2.0 \cite{watanabe2018espnet}.
For the input features, we use 80-band mel-scale spectrogram derived from the speech signal.
The encoder consists of 5-layer pyramidal-BLSTM with subsampling after second and third layers.
The decoder is comprised of 2-layer LSTM with location-aware attention mechanism \cite{chorowski2015attention}.
The target sequence is processed in 5K case-insensitive sub-word units created via unigram byte-pair encoding \cite{shibata1999byte}.
All the LSTM and FC layers have 1024 hidden units each.
Our model is trained for 10 epochs using Adadelta optimizer \cite{zeiler2012adadelta} with learning rate of 1e-8.
Using this baseline acoustic model, we obtain 100-best decoded sentences for each input through hybrid CTC-attention based scoring \cite{hori2017advances} method, and these 100-best lists will be used for rescoring.
Table \ref{tab:wer_oracle} shows the word error rates (WERs) obtained from the baseline model and the oracle WERs, which is the best possible errors of the 100-best lists on the LibriSpeech tasks.
\begin{table}[th]
  \caption{Oracle WERs of the 100-best lists on LibriSpeech}
  \label{tab:wer_oracle}
  \centering
  \begin{tabular}{ccccc}
    \toprule
    \multirow{2}{*}{Method} & \multicolumn{2}{c}{dev} & \multicolumn{2}{c}{test} \\ \cline{2-5}
    &clean  &other  &clean  &other \\
    \midrule
    1-best (baseline) &7.17   &19.79  &7.26   &20.37 \\
    100-best (oracle) &2.85   &12.21  &2.81   &12.85 \\
    \bottomrule
  \end{tabular}
\end{table}

\subsection{Language Model Setups}
\label{sec:lm_set}
The model parameters of our language model (LM) are as follows: $L=3$ for the number of layers, $d=512$ for the dimensions of the model and the embeddings, $h=8$ for the number of head. 2048 hidden units are used in the position-wise feed-forward layers.
We use trainable positional embeddings with supported sequence lengths up to 128 tokens.
We use a \textit{gelu} activation \cite{hendrycks2016bridging} rather than the standard \textit{relu}, following \cite{radford2018improving,devlin2018bert}.
Weight matrix of the softmax layer is shared with the word embedding table.
The word vocabulary is used in three sizes: 10k, 20k and 40k most frequent words.
For a fair comparison, our biSANLM and uniSANLM have the same architecture and parameters except for the vocabulary size.

We train the LMs with the 1.5G normalized text-only data of the official LibriSpeech corpus.
We use Adam optimizer \cite{kingma2014adam} with learning rate of 1e-4, $\beta_1=0.9$, $\beta_2=0.999$. 
We use a dropout probability of 0.1 on all layers.
Batch size is set to 128 for biSANLMs and 64 for uniSANLMs, and all the LMs are trained for 1M iterations.
We confirmed that all our LMs are converged before the 1M training steps.

\section{Results and Discussion}

In this section, we compare the uniSANLM and biSANLM for the $N$-best rescoring on all test sets of the LibriSpeech ASR corpus, in which the test sets are classified as 'clean' or 'other' set based on their difficulties.
We first prepare 100-best hypotheses using our acoustic model (AM), which is referred to as the ``baseline'' recognition system in our experiments.
For rescoring the $100$-best list, the baseline AM is linearly interpolated with one of our language models as in Equation \ref{eq:inter2}.
The interpolation weight is set to a value that achieves the best performance in the development sets.
We find that $\lambda=0.2$ and $0.3$ are the best weights for dev-clean and dev-other sets respectively.
Considering that the dev-other set is more difficult for the acoustic model to recognize, it is reasonable to have larger interpolation weight in dev-other ($\lambda=0.3$) than in dev-clean ($\lambda=0.2$).

\begin{table}[th]
  \caption{WERs for unidirectional and bidirectional SANLMs interpolated with the baseline model on LibriSpeech}
  \label{tab:wer_compare} 
  \centering 
  \begin{tabular}{lrcccc} 
    \toprule 
    \multicolumn{1}{c}{\multirow{2}{*}{Model}} & \multicolumn{1}{c}{\multirow{2}{*}{$|V|$}} & \multicolumn{2}{c}{dev} &  \multicolumn{2}{c}{test} \\ \cline{3-6} 
    & &clean &other   &clean  &other \\ 
    \midrule 
    baseline & &7.17  &19.79   &7.26  &20.37 \\ 
    \midrule 
    \multirow{3}{*}{ + uniSANLM}
        & 10k   &6.09   &17.50  &6.08   &18.33 \\ 
        & 20k   &6.05   &17.48  &6.11   &18.25 \\ 
        & 40k   &6.08   &17.32  &6.11   &18.13 \\
    \midrule 
    \multirow{3}{*}{ + biSANLM}
        & 10k   &5.65   &16.85  &5.69   &17.59 \\ 
        & 20k   &5.57   &16.71  &5.68   &\textbf{17.37} \\ 
        & 40k   &\textbf{5.52}   &\textbf{16.61}  &\textbf{5.65}   &17.44 \\
    \bottomrule
  \end{tabular}
\end{table}
Table \ref{tab:wer_compare} shows rescoring results of the biSANLMs and the uniSANLMs with different test sets and different vocabulary sizes $|V|$.
The WER results show that the biSANLM with our approach is consistently and significantly better than the uniSANLM regardless of the test set and the vocabulary size.

\begin{figure}[th]
    \centering
    \includegraphics[width=0.9\linewidth]{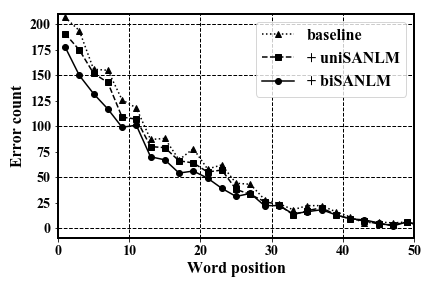}
    \caption{Error count by word position}
    \label{fig:analysis}
\end{figure}
To see where the ``word error'' occurs, we analyze the position of the misrecognized words.
Figure \ref{fig:analysis} shows the total number of the misrecognized words for each model according to the position of the final hypotheses.
It can be seen that the biSANLM is more robust than the uniSANLM at the earlier position ($<30$) of a sentence.
At the latter position ($>30$) of a long sentence, however, the gap between the two LMs is reduced.
This fact shows that, although the uniSANLM also performs well, our biSANLM is more effective particularly when a recognized sentence is short or a misrecognized word is at the beginning of the sentence.
In this analysis, we use $|V|=40$k for both LMs, and the choice of vocabulary size does not affect the tendency.

Finally, we conduct linear interpolation of the two LMs for further improvements: 
\begin{equation*}
    \text{score}_\text{LM}= (1-\alpha)\cdot\text{score}_{\text{uniLM}} + \alpha\cdot\text{score}_{\text{biLM}},
\end{equation*}
where the $\text{score}_\text{LM}$ is used in Equation \ref{eq:inter2}. We find $\alpha=1$ shows the best performances on all test sets, which means only the biSANLM is used for interpolation (log-linear interpolation of the two LMs shows the same phenomenon).
Contrary to our first expectation, the biSANLM and the uniSANLM do not complement each other in our experiments.

Consequently, all experimental results demonstrate that our sentence scoring method using the biSANLM is almost strictly better than the traditional method using uniSANLM for the $N$-best list rescoring.
As far as we know, this is the first study that the bidirectional language model significantly and consistently outperforms the unidirectional language model for speech recognition.

\section{Conclusion}
In this paper, we propose a novel sentence scoring method that uses a biLM for rescoring in ASR.
We used the biLM to predict the probability of the masked word, and thus made the model enable to capture the interactions between the past and the future words.
Experimental results on the LibriSpeech ASR tasks demonstrated that the proposed sentence scoring with our biLM significantly and consistently outperforms the conventional uniLM for rescoring the $N$-best list.
In addition, we confirmed that the biLM is more robust than the uniLM especially when a recognized sentence is short or the earlier part of the sentence is misrecognized.


\bibliographystyle{IEEEtran}
\bibliography{mybib}

\end{document}